\documentclass{article}

\usepackage{PRIMEarxiv}

\errorcontextlines\maxdimen

\usepackage[utf8]{inputenc} 
\usepackage[T1]{fontenc}    
\usepackage{url}            
\usepackage{booktabs}       
\usepackage{nicefrac}       
\usepackage{microtype}      
\usepackage{fancyhdr}       
\usepackage{graphicx}
\usepackage{xcolor}
\usepackage{threeparttable}
\usepackage{booktabs}
\usepackage{multirow}
\usepackage{tabularx}
\usepackage{arydshln}
\usepackage{subcaption}
\usepackage{cite}

\usepackage{lineno,hyperref}
\hypersetup{colorlinks, citecolor=blue, linkcolor=red, urlcolor=blue}
\usepackage{amsmath,amsfonts,amssymb}
\usepackage{graphics}
\usepackage{bm}
\usepackage{multirow}
\usepackage{algorithm}
\usepackage{algpseudocode}

\usepackage{mathtools}
\usepackage{siunitx}
\DeclarePairedDelimiterXPP\BigOSI[2]%
  {\mathcal{O}}{(}{)}{}%
  {\SI{#1}{#2}}

\newcommand\blfootnote[1]{%
  \begingroup
  \renewcommand\thefootnote{}\footnote{#1}%
  \addtocounter{footnote}{-1}%
  \endgroup
}
  
\title{\emph{G\textsuperscript{2}TR}: \underline{G}eneralized \underline{G}rounded \underline{T}emporal \underline{R}easoning for Robot Instruction Following by Combining Large Pre-trained Models}

\author{Riya Arora\footnote{\small $^{\{*,+\}}$denotes equal contribution as joint authors} \\
  Indian Institute of Technology Delhi\\
  \texttt{arorariya2906@gmail.com} \\
  \And
  Niveditha Narendranath$^{*}$\\
  Indian Institute of Technology Delhi\\
  \texttt{nivinath7@gmail.com} \\
  \And
  Aman Tambi\footnote{\small $^{\{*,+\}}$denotes equal contribution as joint authors} \\
  Indian Institute of Technology Delhi\\
  \texttt{a.amantambi@gmail.com} \\
  \And
  Sandeep S. Zachariah$^{\dagger}$ \\
  Indian Institute of Technology Delhi\\
  \texttt{sandeep.s.zac@gmail.com}
  \And
  Souvik Chakraborty \\
  Indian Institute of Technology Delhi\\
  \texttt{souvik@am.iitd.ac.in} \\
  \And
  Rohan Paul \\
  Indian Institute of Technology Delhi\\
  \texttt{rohan@cse.iitd.ac.in} \\
}

\begin{document}

\maketitle

\begin{abstract}
Consider the scenario where a human cleans a table and a robot observing the scene is instructed with the task \textit{``Remove the cloth using which I wiped the table"}. Instruction following with temporal reasoning requires the robot to identify the relevant past object interaction, ground the object of interest in the present scene, and execute the task according to the human's instruction. 
Directly grounding utterances referencing past interactions to grounded objects is challenging due to the multi-hop nature of references to past interactions and large space of object groundings in a video stream observing the robot's workspace. Our key insight is to factor the temporal reasoning task as (i) estimating the video interval associated with event reference, (ii) performing spatial reasoning over the interaction frames to infer the intended object  (iii) semantically track the object's location till the current scene to enable future robot interactions. Our approach leverages existing large pre-trained models (which possess inherent generalization capabilities) and combines them appropriately for temporal grounding tasks. Evaluation on a video-language corpus acquired with a robot manipulator displaying rich temporal interactions in spatially-complex scenes displays an average accuracy of \textbf{70.10\%}. The dataset, code, and videos are available at \urlstyle{rm}
\url{https://reail-iitdelhi.github.io/temporalreasoning.github.io/}

\end{abstract}


\section{Introduction}
\blfootnote{\small $^{\{*,\dagger\}}$denotes equal contribution as joint authors}
Imagine an assistive robot deployed in a human co-habit-ed space. The robot continually accrues observations of rich human-object interactions occurring in its workspace, e.g., objects may be placed, moved, re-oriented, packed away etc. In such setting, our goal is to interpret and follow instructions that require contextual reasoning over time. For example, \emph{``robot, hand me the cup that was last poured into"} , \emph{``clear away the object used for wiping"} or \emph{``where did I keep my spectacles?"} require the robot to either bring the object or to gesturing/pointing to the location of the observed object or the containing object (if occluded). 

Grounding such instructions requires the robot to foviate when interactions occurred and perform temporally extended (multi-hop) reasoning. Further, for the robot to plan interactions with the object, the intended object must be grounded in the metric space (e.g., as image bounding boxes) and knowledge of its location must be propagated to the current time step accounting for further interactions such as re-positioning or placing inside another object. 
Performing such grounded temporal reasoning is challenging  due to the expansive space of  human-object interactions (moving, re-orienting or packing), need for multi-hop reasoning over interaction sequences and grounding the current object location under possibility (possibly under occlusion in case of containment). 

\begin{figure}[t!]
  \centering
\includegraphics[width=\columnwidth, trim=0cm 6cm 0cm 6cm]{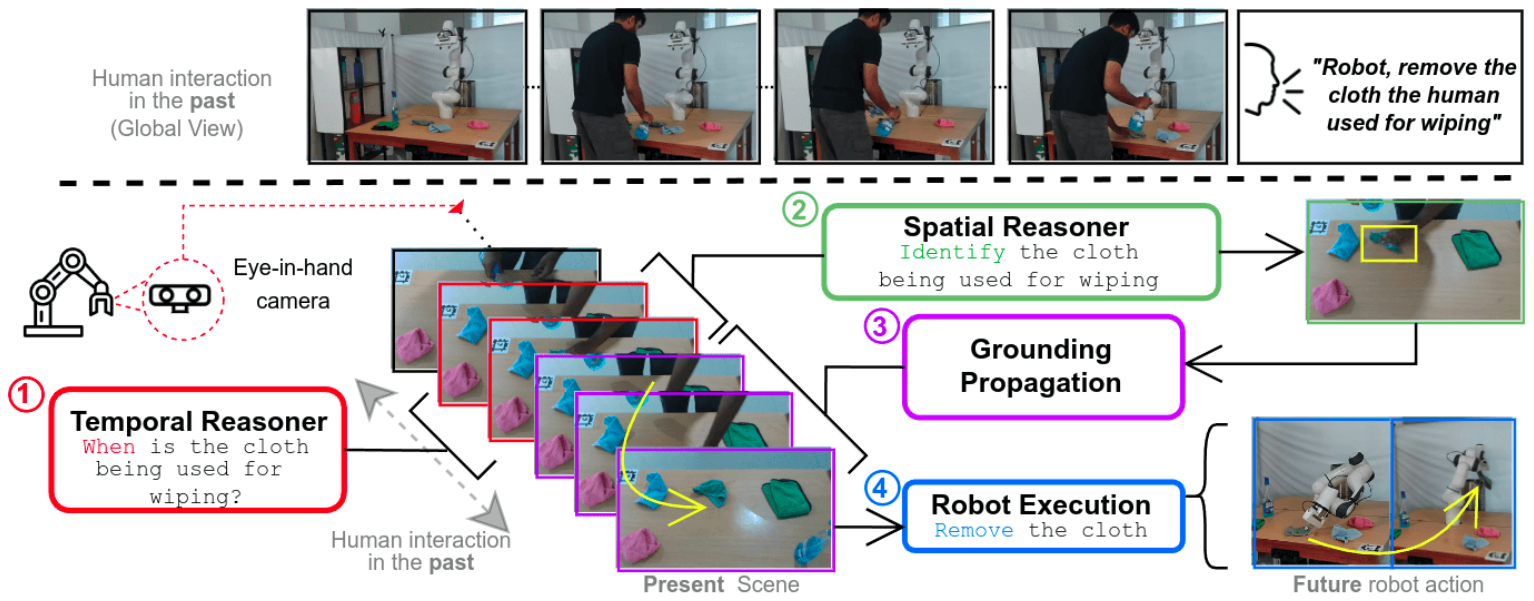}
  \caption{ 
  \textbf{Method Overview.} We propose a method for following instructions involving reasoning over past interactions to determine future robot actions. We combine large pre-trained models (possessing inherent generalization capacity) for (i) temporally localising events (via video-LLM), (ii) spatially grounding intended object in interaction frames (via visual-QA aided with visual prompts) and (iii) propagating knowledge of its location (via semantic segmentation) for future robot interaction.\label{Fig_1:rw} 
}
  \vspace{-15pt}
\end{figure}

Approaches for temporal grounding exhuastively extract factual knowledge from the past \cite{robomem} or  learn to infer \cite{robomem,temporalgg,barbu2012video} instruction-relevant facts from past observations to ground instructions. The need to specify the space of interactions \emph{a-priori} and the nature of training makes the approaches difficult to scale to a vast (open) set of interactions as well as complex scenes. The recent emergence of large pre-trained models associate free-form language instructions with visual or video observations. However, their focus on visual/question answering does not explicitly model the need for explicit grounding of the object in the image space
and hence cannot be used directly to execute future robot actions. 

This paper presents a method for grounded temporal reasoning that builds on existing pre-trained visual/video-language models with inherent open-set generalization capacity which are then coupled in a novel way. As shown in Fig. \ref{Fig_1:rw}, our approach factors inference as: (i) temporal reasoning associating declarative aspects of the instructions with image sequences (ii) grounding the inferred object into image sequences using reference grounding models and (iii) propagating semantic knowledge of the grounded object from referred event to the current world state (via semantic segmentation and common sense reasoning) enabling future robot interactions. Evaluation on a real-world dataset collected using a robot manipulator (with an eye-in-hand camera) demonstrates an average grounding accuracy of $70.10\%$ improving over alternatives by $\sim26.62\%$. We release code, dataset and videos to the research community. .

\section{Related Works}\label{RW}
\textbf{Robot Instruction Following.} A rich body of work focuses on grounding/associating language commands with intended robot action,  \cite{kollar2013generalized,chung2015performance,misra2016tell,blukis2020few, mon2024enabling}. A key limitation of these systems is their inability to refer to or reason over past interactions, which would be required for tasks such as ``Robot, pick the object that was dropped by the person'' or ``Robot, help the person who just fell''. Through our framework, we aim to inculcate temporal reasoning abilities in such robot-instruction following systems so that the aforementioned tasks can be realized. A related set of approaches, \cite{idrees2020robomem,temporalgg,tellex2020robots,idreestowards} focus on sequential instruction following over an extended time period. Although these methodologies provide efficient solutions for dealing with temporal and linguistic complexities, they are limited to supervised settings, restricting the objects to a fixed set. There are approaches, \cite{po_1, po_2, po_3}, focussing on Partial Observability in the spatial domain; however, they do not tackle Partial Observability in the temporal domain. Generalization being the impediment, our approach involves leveraging the capabilities of recently developed Foundational Models, especially their competence in zero-shot learning. This enables us to work with a diverse and representative dataset unrestrained by any hand-crafted features.

\textbf{Vision-language Reasoning.} Complementary efforts, \cite{tr_vid_1, tr_vid_2, tr_vid_3}, explore temporal reasoning in videos, including a few, \cite{barbu2012video,siddharth2014seeing,barrett2015saying}, specifically focusing on retrieving relevant semantic relations over time and utilizing it to perform temporal grounding in videos. These aid in solving complex linguistic intricacies as well as generalized temporal event localization, but their applications have not yet been extended to actual implementation on robots in the real-world. To fill this gap, our approach on the other hand, incorporates a motion planning module that ensures the intended action is executed by the robot. The recently emerged multi-modal, pre-trained, Large Vision Language Models, \cite{hong2024cogvlm2,videochatcaptioner,videollava,videochatgpt,cogvlm,gpt4,zhang2023video,huang2024lita,videochat2,liu2024tempcompass,tang2023video}, are capable of performing reasoning over images and videos given queries in free-form text. Their abilities are manifold and include interpreting natural language to enable visual understanding, visual question-answering as well as visual captioning. A different set of foundation models are the phrase-grounding models, \cite{li2022grounded, chen2019uniter, kamath2021mdetr, cogvlm, dino}, which possess grounding abilities in visual modality. Complications arise because although temporal and spatial reasoning is done in a zero-shot nature over diverse data using these models, there is no one single model that integrates all the aforementioned tasks into one cohesive and sophisticated system.

\section{Problem statement}\label{PS}
We consider a robot manipulator operating in a table top setting. The robot is capable of acquiring visual observations, taking language instructions and has the ability to take several actions in its action space such as picking, pointing, removing etc. Such manipulation behaviours and a planner is assumed known to the robot. Given that some human-object interactions took place in the environment and an input language instruction was given to a robot, our goal is to program the robot to reason over accrued visual observations and execute the instruction by taking appropriate actions.  
 
Formally, given an input language instruction ($\Lambda$) and past observations ($I_{0:t}$) capturing human-object interactions in the workspace, the robot's task is to infer the intended object and ground the object $o$ in the current scene image. We assume the grounding in the form of a 2D bounding box $\beta_{o} \in R^{4}$.  
\begin{equation}\label{eq:1 - Temporal Grounding}
    \text{TemporalGrounding} (I_{0:t}, \Lambda) \rightarrow \beta_{o} 
\end{equation}

We assume known camera calibration allowing the robot to back-project 2D bounding boxes into 3D world model enabling the motion planner to plan its trajectory for future interaction with the grounded object. 

Grounding language instructions (which embody reasoning over observations over time) directly to intended objects in the scene (expressed as bounding boxes) is challenging. Hence, we adopt a factored approach of performing temporal reasoning over image sequences to ground the interaction and then perform spatial reasoning and referring expression grounding over a subset of images (containing the interaction) to resolve the intended object. Finally, the object is tracked till the current frame to enable future robot interactions with the object. Further, we consider generalization to the open set setting and assume that the robot does not have \emph{a-priori} knowledge of which object types may be present in the scene and how the human would interact with them over time (picking, placing, re-orienting, wiping, pushing etc.). Hence, we leverage foundation model and present a novel combination for grounded temporal reasoning. We detail the pipeline in the next section.

\section{Grounded Temporal Reasoning}\label{Framework}
We address the temporal reasoning task by grounding the requisite interaction in the input-video for the candidate interval extraction, and then pinpointing the intended object through fine-grained spatial reasoning within this interval for grounding. Our framework is comprises three components: (1) Temporal-Parser (TP) (2) Candidate interval estimation via Event-Localizer (EL), and (3) Grounding object of interest via Target-Detector (TD)
followed by grounding propagation via semantic tracking.

\begin{figure*}[htbp]\label{Fig 2: Approach Overview}
\centering
\includegraphics[width=\columnwidth, trim=0cm 10.7cm 0cm 11cm]{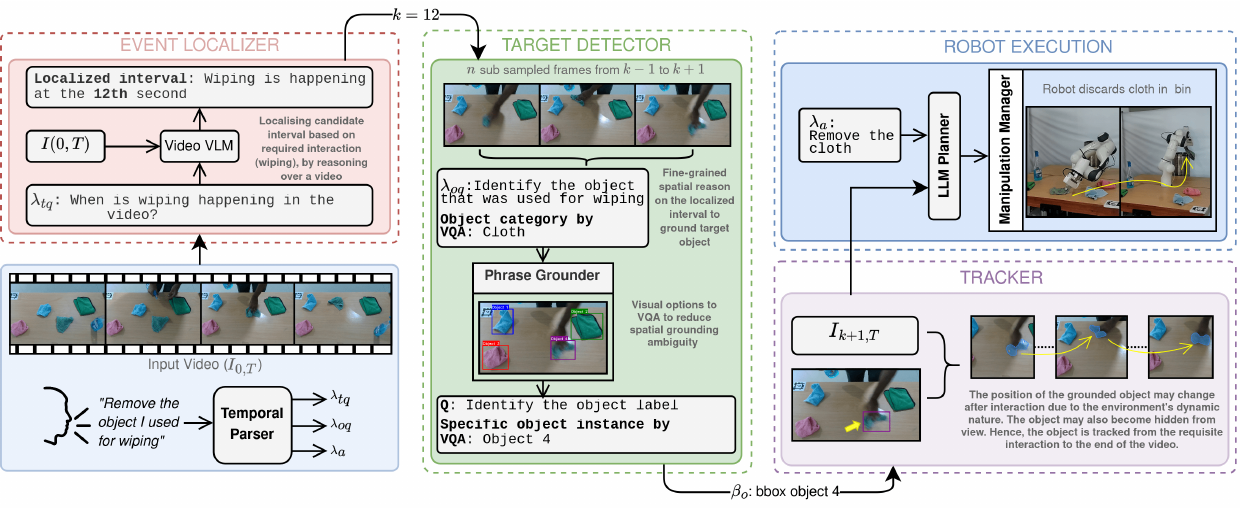}
\caption{\textbf{Pipeline Overview.} Event Localization processes the input video based on a temporal query. Using the output timestamp, a frame interval is generated. The Target Detector identifies the precise object through visual prompting, and the Tracker follows it using bounding boxes. The robot then executes the action.}
\label{fig:pipeline_diagram}
\end{figure*}

\subsection{Temporal Parsing}
This module realizes the need for reasoning over both the past and language-referenced object-interaction by generating a temporal question ($\lambda_{\text{tq}}$) for the Event Localizer module and object-identification question ($\lambda_{\text{oq}}$) for the Target Detector module. So, for the instruction, \textit{``Robot, remove the cloth used for wiping"}, the parser gives, \textit{``When was the cloth being used for wiping"} (a temporal question) and \textit{``Identify the cloth that was used for wiping"} (an object identification question). Additionally, the temporal parser extracts the action ($\lambda_{\text{a}}$) that the robot needs to perform on the target object (here, \textit{``Remove"}). This is accomplished through in-context learning by providing a large language model (LLM) with input instruction. (\(\Lambda\)). 
\begin{equation}\label{eq:2 - Temporal Parser}
    \text{TemporalParser} (\Lambda) \rightarrow \{\lambda_{\text{tq}}, \lambda_{\text{oq}}, \lambda_{\text{a}}\}
\end{equation}

\subsection{Estimating Candidate Interval}
This module performs temporal reasoning on accrued past observations to identify the likely interval of the required interaction.The interval in the video where the language-referenced interaction occurs is extracted using a video-understanding Visual Language Model. The video (\(I\textsubscript{0:T} \in \mathbb{R}^{T \times H \times W}\)) and the temporal question ($\lambda_{\text{tq}}$) is taken as input, and the time instant k, (k$\in{\{0, 1, .., T\}}$) when the interaction occured is obtained. If the temporal question \textit{“Robot, when is the object being used for wiping?”}, this module outputs, \textit{``Wiping is taking place at the \textbf{$12^{th}$} second."}

\begin{equation}\label{eq:3 - Event Localizer}
    \text{EventLocalizer} (I\textsubscript{0:T}, \lambda_{\text{tq}}) \rightarrow k
\end{equation}
Once the time instance \(k\) is identified, an interval is constructed, ranging from $(k-1)^{th}$ to $(k+1)^{th}$ second. Within this interval, we uniformly sub-sample a set of $n \in \mathbb{Z}$ frames ($S_{n} = \{I_{1},\ldots, I_{n}\}$) for the next module.\footnote{With a video frame rate of 30 fps, frames are selected from the start of the \((k-1)^{\text{th}}\) second to the start of the \((k+1)^{\text{th}}\) second, resulting in a total of 60 frames. Out of these, 15 frames are uniformly sub-sampled for Target Detector Module in our experiments.}

\begin{equation}\label{eq:4 - Interval Constructor}
    \text{IntervalConstructor} (I\textsubscript{0:T}, k) \rightarrow S_{n} 
\end{equation}

\subsection{Grounding Object of Interest}
The purpose of this module is to perform fine-grained spatial reasoning and ground the target-object involved in the interaction. A set of frames from previous module ($S_{n} = \{I_{1},\ldots, I_{n}\}$) and object-identification question (\(\lambda_{\text{oq}}\)) are sent as input. The output is the grounded coordinates in 2-D ($\beta_o$) of the target object in the last frame of this interval. This is done via a three step process as shown in \ref{fig:pipeline_diagram}. First the target-object \textit{class} (``cloth") is identified by an image-understanding VLM, then a Phrase Grounder is used for grounding objects belonging to this class on the last image ($I_{n}$) in interval. The VLM is re-prompted using visual prompting strategies, as in \cite{visual-prompt1,visual-prompt2, visual-prompt3}, from the Grounder, which then selects the target object from these visual objects/ options (``Object 4").

\begin{equation}
    \label{eq:5 - Target Detector}
    \text{TargetDetector} (S_{n},\lambda_{\text{oq}}) \rightarrow \beta_o
\end{equation}

\subsection{Grounding Propagation via Semantic Tracking}
Additional interactions with the object after the language-referred interactions can lead to the object being in different locations (including contained within or occluded by another objects).
Hence, we need reasoning over the grounded object state and location from the time of past interaction till the current world state for future robot manipulation.

\begin{figure}[t]\label{Fig 3: Grounding Propagation}
\centering
\includegraphics[width=\columnwidth, trim=0cm 8.7cm 0cm 9cm]{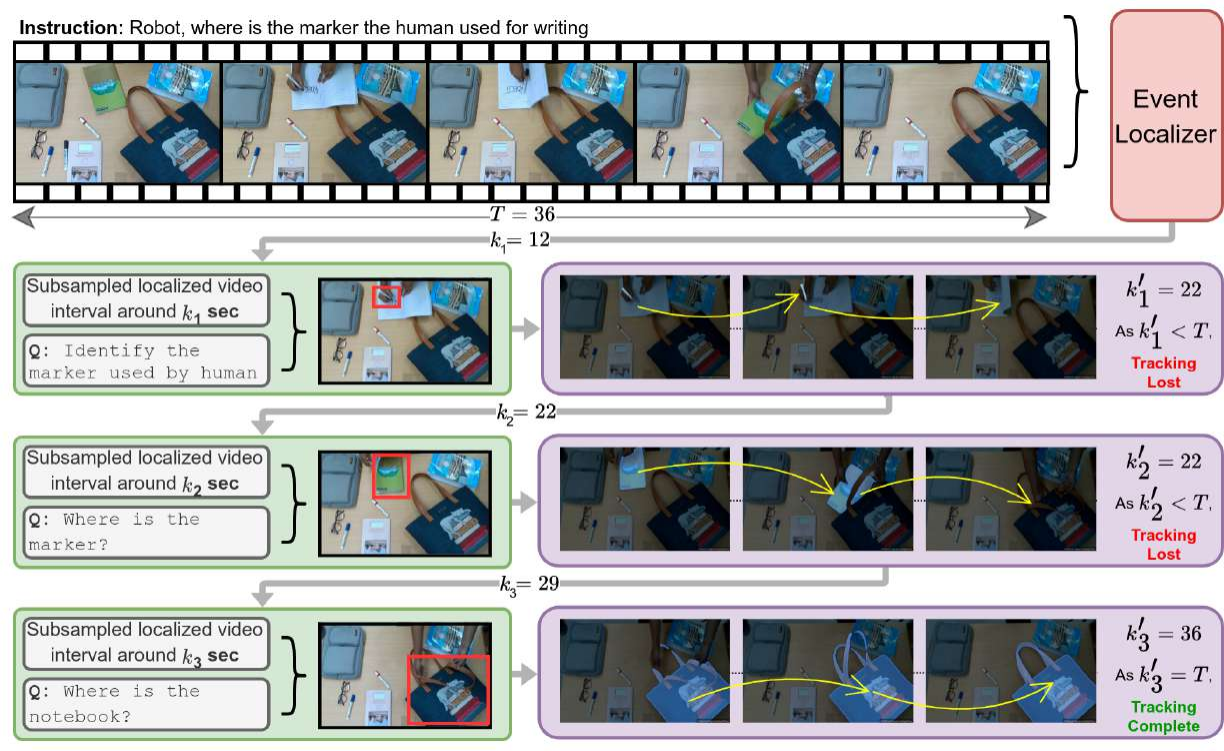}
\caption{\textbf{Grounding Propagation.} Semantic tracking via a vision foundation model is used to propagate the location of the grounded object from the end of the language-referred interaction to the the current world state. Tracking failure (when the object goes out of view) triggers language re-prompting to estimate where the object is and to  track the containing object till the current world state.}
\end{figure}

We address this issue by introducing (i) tracking the object till present scene, and (ii) changing the target object from the intended object to one that occludes it, if occlusion occurs.
\begin{equation}\label{eq:6 - Tracker}
\text{Tracker}(I_{\text{k+1:T}}, \beta_o^{k+1}) \rightarrow \beta_o^{k'}, k' = T
\end{equation}
In real-life scenarios, the object we want to locate is often partially or completely concealed, motivating us to devise a solution for cases within the domain of partial observability. We handle such cases through an iterative process where the Target Detector and the Tracker are invoked repeatedly until the Tracker returns the grounded coordinates of the intended object corresponding to the last second.

Let $i = 1, \ldots, N$ where $N$ is the total number of iterations (or the instances when target object goes out of view). 

\begin{equation}\label{eq:7 - PO Tracking}
\text{Tracker}(I_{k_{i-1}+1:T}, \beta_o^{k_{i-1}}) \rightarrow \beta_o^{k'_{i-1}}
\end{equation}
where $k'_{i-1} < T $ and represents the time instance till which the target object was tracked. As and when the target object goes out of view, an interval ranging from $k'_{i-1}-1$ to $k'_{i-1}+1$ is constructed and a set of $n$ frames $(S_{n_{i}})$ is once again sub-sampled. The target detector module is then re-prompted to identify the occluding object and invoked again as described in eqn \ref{eq:5 - Target Detector}.

\begin{equation}\label{eq:8 - PO Target Detection}
\text{TargetDetector}(S_{n_{i}}, \lambda_{oq})\rightarrow \beta_o^{k_i}
\end{equation}

This iterative cycle keeps on occuring till the target object is detected in the present world scene of the robot, so that it can take the requisite actions in the future.

\section{Experiment Setup}\label{ES}
\textbf{Evaluation Corpus.}  To assess the robot's ability to perform grounded temporal reasoning, we form an evaluation corpus of $155$ video-instruction pairs. The data set was collected in a table-top setting with a Franka Emika Robot Manipulator observing the scene via a eye-in-hand RGB-D camera. A total of $15$ objects representing common household and healthcare items such as cups, bottles, medicines, fruits, notebooks, markers, handkerchiefs etc. were used. A total of $8$ participants\footnote{The study followed institute human-subject evaluation guidelines and took informed consent from the participants} performed interactions such as pouring, placing, picking, stacking, replacing, wiping, dropping, repositioning, swapping etc. with $3-6$ objects in each scene. The human participants also provided natural language instructions for the robot referencing to interactions and objects in the preceding interactions in the workspace. This resulted in a corpus of $155$ video instruction pairs (each between $6-30$ seconds). The instructions and interactions expressed natural diversity of spatial and temporal reasoning complexity. For detailed analysis, the evaluation corpus is bifurcated as: (i) single/multi hop temporal reasoning $(56:99)$ (ii) simple/complex spatial grounding $(98:57)$, (iii) single/multiple interactions $(62:93)$ and (iv) partial or full observation of the referred object in the video $(36:119)$, see (Fig. 3). The ground truth bounding boxes corresponding to the intended object for each instruction-world pair were assigned 
hand-engineering prompts for a phrase grounding model (CogVLM \cite{cogvlm}) by a human annotator. Further, each bounding box label was verified by at least two annotators.\\ 

\begin{figure}[!t]\label{Fig 4 : Evaluation Corpus}
\centering
\includegraphics[width=\columnwidth, trim=0cm 9.5cm 0cm 10cm]{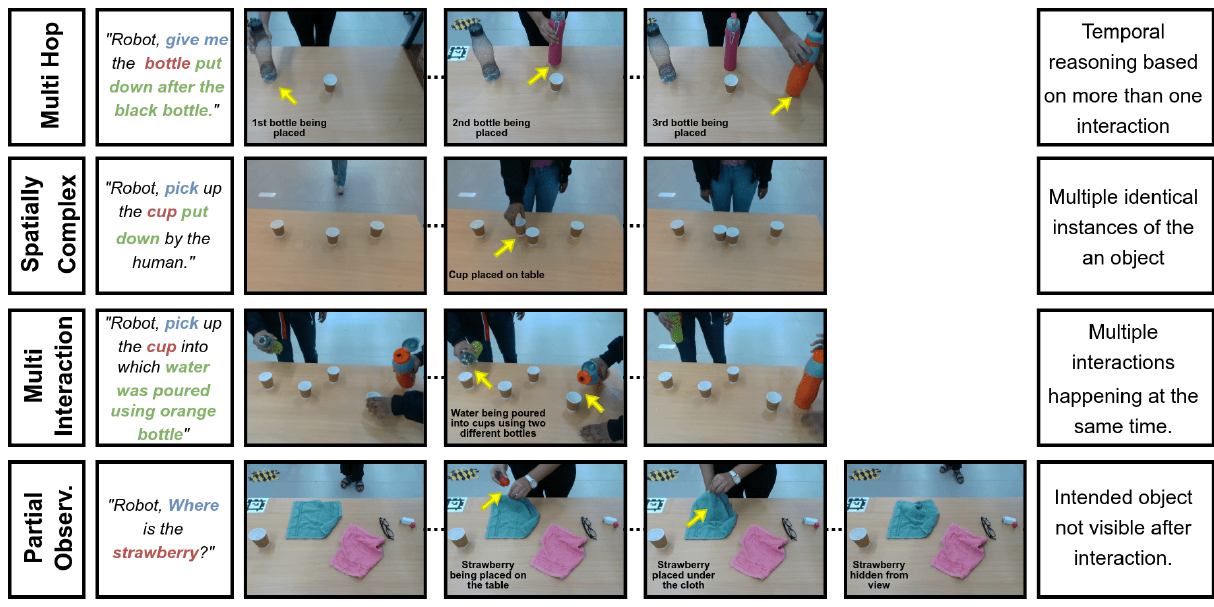}
\caption{\textbf{Evaluation Corpus Bifurcations.} The evaluation corpus is bifurcated for analysis as per the complexity of temporal, linguistic and spatial reasoning required for grounding. Figure shows representative examples.}
\end{figure}

\textbf{Models.} The proposed model was evaluated with two baseline approaches reflecting alternative ways to factor the temporal reasoning problem. Each translates to a different architecture for combining pre-trained models. Note that we only include open-set models in our evaluation for their generality and do not evaluate approaches such as that are restricted to specific object and interaction classes. 

\textbf{Proposed Model (G$^2$TR).} This model was realized with a video-understanding VLM (CogVLM2-video\cite{hong2024cogvlm2}) for candidate interval identification, followed by target detection and grounding, achieved using a Vision Language Model (GPT-4o \cite{gpt4}) and a Phrase Grounder (GroundingDino1.5-pro \cite{dino}). This was further integrated with a tracker module (SAM-2 \cite{ravi2024sam2}) to account for dynamism in robot's environment.

\textbf{Direct Temporal Visual Grounding (DTVG).} This approach involves directly using video-understanding Visual Language Model (VLM) for both temporal and spatial reasoning followed by a Phrase Grounder for intended-object grounding. So, for instructions like ``Robot, remove the cloth which was used for wiping", the video-understanding VLM would be prompted to identify the cloth used for wiping and is expected to return the intended object and its description, eg `green cloth on the left'. This linguistic description is then used by the grounder module to ground the object of interest i.e. left green cloth.

\begin{figure}[h!]\label{Fig 5.1 : Alternate Approach (I)}
\centering
\includegraphics[width=\columnwidth, trim=0cm 12.5cm 0cm 13cm]{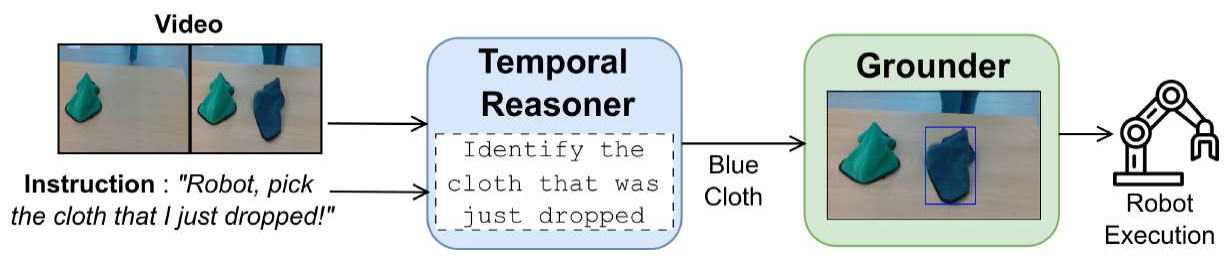}
\caption{Alternate Approach - Direct Temporal Visual Grounding}
\end{figure}
 
\textbf{Refined Temporal Visual Grounding (RTVG).} This approach is similar to the previous approach (DTVG), the only difference being that it utilizes VQA (Visual Question Answering) performed by a VLM, to iteratively enhance the spatial linguistic description of the intended object thereby aiding grounder module. So, for the aformentioned instruction, if the video-understanding VLM returns just `cloth', it would be re-prompted till it gives enough description (`green cloth on the left') to ground the intended object.\\

\begin{figure}[h!]\label{Fig 5.2 : Alternate Approach (II)}
\centering
\includegraphics[width=\columnwidth, trim=0cm 13cm 0cm 15cm]{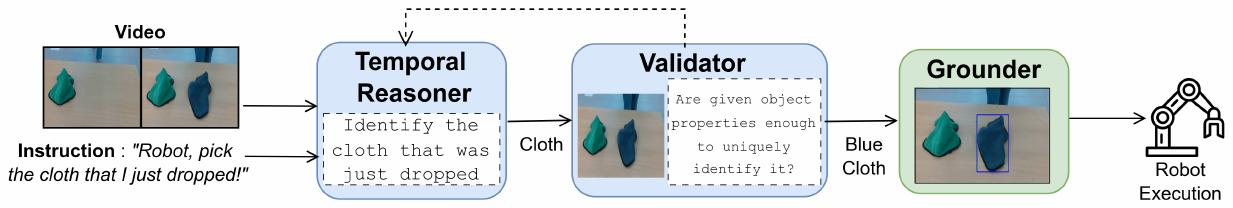}
\caption{Alternate Approach - Refined Temporal Visual Grounding}
\end{figure}

\textbf{Evaluation Metrics. } Models were evaluated in terms of accuracy of inferring the object as a result of contextual temporal reasoning. The grounding accuracy is computed via Intersection over Union (IoU) between the bounding box for the intended object obtained from model output and the corresponding ground truth (thresholded at $0.7$). Further, for component-wise evaluations, correct or incorrect ground labels were assigned via human inspection for intermediate outputs. For example, for the event localization module, the human assessed if the input utterance such as ``Robot, which object was placed on the table?" was correct for a given video and further, if the event was localized correctly in the frame indices provided. Similarly, ground truth was also assigned for the target object identification and tracking modules.

\section{Results}\label{Results}
\begin{table}[h!]\label{Table 1 : Main Results}
\centering
\renewcommand{\arraystretch}{1.6}
\scalebox{0.75}{
\begin{tabularx}{1.295\textwidth}{|c|c||c|c||c|c||c|c||c|c||c|c|}
\hline
\textbf{Approach} & \textbf{Acc. (\%) \scriptsize\(\uparrow\)} & \multicolumn{2}{c||}{\textbf{Reasoning Depth}} & \multicolumn{2}{c||}{\textbf{Spatial Complexity}} & \multicolumn{2}{c||}{\textbf{Observability}} & \multicolumn{2}{c||}{\textbf{Interaction-Complexity}} \\
\cline{3-10}
& & \textbf{SH (\%) \scriptsize\(\uparrow\)} & \textbf{MH (\%) \scriptsize\(\uparrow\)} & \textbf{SS (\%) \scriptsize\(\uparrow\)} & \textbf{SC (\%) \scriptsize\(\uparrow\)} & \textbf{CO (\%) \scriptsize\(\uparrow\)} & \textbf{PO (\%) \scriptsize\(\uparrow\)} & \textbf{SI (\%) \scriptsize\(\uparrow\)} & \textbf{MI (\%)} \\
\hline
$\text{G}^{2}\text{TR}$ & \textbf{70.10$\pm$2.7} & \textbf{73.73$\pm$2.18} & \textbf{64.28$\pm$5.05} & \textbf{70.41$\pm$4.16} & \textbf{70.17$\pm$1.43} & \textbf{70.59$\pm$3.14} & \textbf{69.44$\pm$4.54} & \textbf{78.49$\pm$3.80} & \textbf{64.87$\pm$3.32} \\
\hline
DTVG & 42.15$\pm$0.80 & 44.44$\pm$0.82 & 38.70$\pm$0.84 & 54.08$\pm$0.83 & 22.22$\pm$2.18 & 49.30$\pm$0.79 & 19.44$\pm$0.00 & 47.85$\pm$2.01 & 38.70$\pm$1.52 \\
\hline
RTVG & 44.81$\pm$4.78 & 46.46$\pm$3.78 & 41.67$\pm$6.89 & 55.44$\pm$3.94 & 26.31$\pm$5.73 & 48.46$\pm$5.15 & 32.41$\pm$2.62 & 49.46$\pm$3.31 & 41.58$\pm$5.36 \\
\hline
\end{tabularx}
}
\caption*{\textnormal{\textbf{Table 1.} Accuracy of all approaches: our pipeline (G$^{2}$TR), Direct Temporal Visual Grounding (DTVG), and Refine Temporal Visual Grounding (RTVG), across all sub-categories of dataset bifurcations - Reasoning Depth (SH: Single-Hop, MH: Multi-Hop), Spatial Complexity (SS: Spatially-Simple, SC: Spatially-Complex), Observability (CO: Completely-Observable, PO: Partially-Observable), Interaction-Complexity (Single-Interaction(SI), Multi-Interactions(MI)) }}
\label{table:comparison}
\end{table}

\begin{figure*}[h!]
    \centering
    \includegraphics[width=\columnwidth, trim=0cm 9.5cm 0cm 9cm]{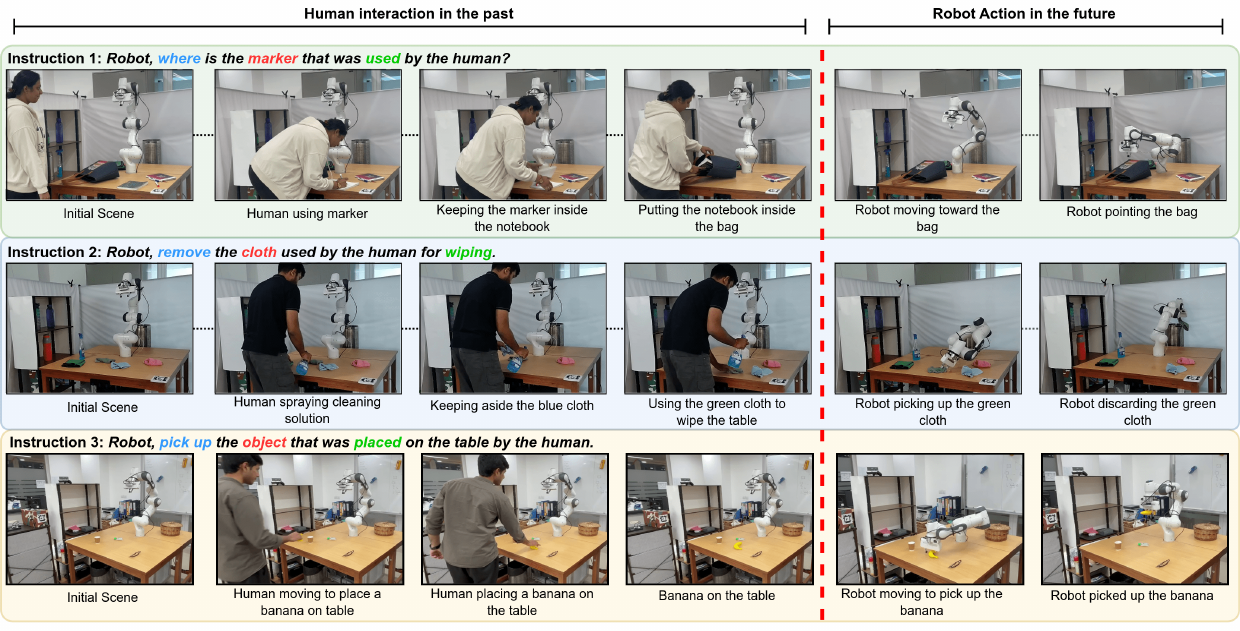}
    \caption{\textbf{Robot Experiments}. The figure illustrates the execution of our pipeline by the robot across various categories from
the dataset. \textbf{Task 1}: robot handling a partially observable object scenario twice. \textbf{Task 2}: robot performing spatial reasoning
due to the presence of similar objects on the table. \textbf{Task 3}: robot tracking objects due to repositioning after human interaction.}
    \label{fig:real-world}
\end{figure*}

\subsection{Evaluation of Proposed Approach}
G$^2$TR has an overall average accuracy of $70.10\%$ which is about $26\%$ better as compared to its nearest competitor, see Table 1. Among different sub-categorizations of the tasks, G$^2$TR yield the best result for single interaction and single hop reasoning, and the performance declines for multi-interaction and multi-hop reasoning scenarios. This is because of the video-understanding model's lack of training for complex situations requiring reasoning over multiple interactions and their relationships. 
Fig.~\ref{fig:real-world} shows the robot successfully following instructions referring to interactions in the past. Experiments are performed on a 7DoF Franka Panda Emika robot manipulator with a parallel jaw gripper. The robot observes a human interact naturally with objects via an Intel Realsense D435i RGBD camera mounted above the end-effector in an eye-in-hand configuration. The experiments leverage a perception pipeline that constructs an object-centric scene graph by back-projecting 2D bounding boxes (output from the temporal reasoning process) into 3D given calibrated camera intrinsics. Grasping synthesis is via \cite{fang2023anygrasp} and actions are sequenced using an LLM-based planner.
Fig.~\ref{fig:component-failure}(a) shows that the bottleneck of G$^2$TR is the Event Localizer's difficulty in precisely identifying the time instant of the language-referenced interaction.  The next bottleneck occurs in the Target Detector, which struggles with selection of incorrect object class, insufficient visual prompts, or incorrect selection of object label due to ambiguity in visual options. The Tracker occasionally begins tracking the wrong object, when it appears identical to the correct one. Fig.~\ref{fig:component-failure}(b).

While our approach performs well in complex scenarios, it struggles with rapid human-object interactions and single actions involving multiple objects, such as replacing or stacking. 
Another limitation of our pipeline is its tendency to perform reasoning based on linguistic cues rather than visual evidence. For instance, when tasked with identifying the object into which water was poured, it may incorrectly output a cup, even if the water was poured into a different container. 
G$^2$TR takes an average of $46.98\pm1.48$ seconds, with most time spent on target detection ($32.26\pm1.11$ s). Under partial observability, runtime nearly doubles to $88.57\pm7.26$ seconds due to iterative detection and tracking, limiting zero-shot generalization and indicating the need for further improvements.
\subsection{Evaluation of Alternative Approaches} 
We also compare the proposed model with alternative approaches (DTVG and RTVG), as shown in Table 1. G$^2$TR achieves a 26\% higher overall average grounding accuracy. The results indicate that reasoning over the entire video stream using an video-understanding VLM is less effective than combining the video-understanding VLM for coarse temporal tasks with another model for fine-grained tasks. Since Multi-Hop (MH) and Multi-Interaction (MI) reasoning are inherently temporal, G$^2$TR is, respectively, 24.10\% and 24.73\% more accurate. 
In spatially complex cases, DTVG and RTVG perform poorly due to linguistic ambiguity in spatial properties. G$^2$TR improves performance by 45.90\% through visual prompting, enabling spatial reasoning in the visual domain. Similarly, DTVG and RTVG struggle with partial observability, likely due to a lack of training on such examples. G$^2$TR overcomes this by using a separate module to track occluding objects, improving performance by 41.01\%.

\begin{figure}[h!]
    \centering
    \includegraphics[width=0.9\columnwidth, trim=0cm 9.5cm 0cm 9.5cm]{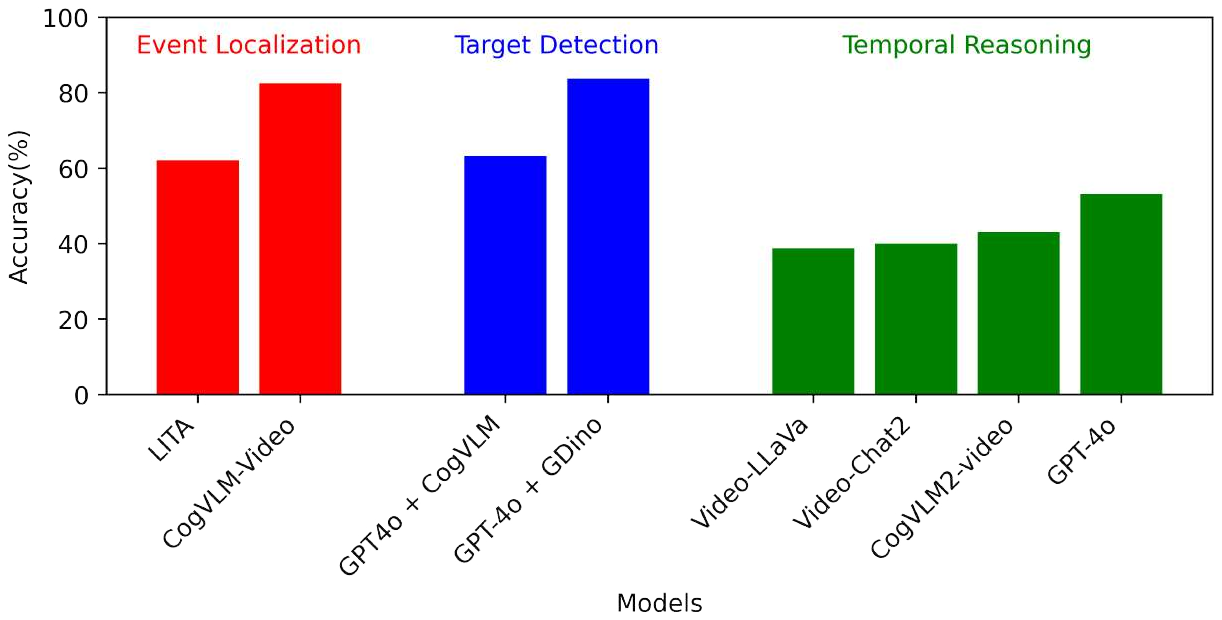}
    \caption{Task-wise Model Analysis}
    \label{fig:model_analysis}
    \vspace{-1.5em}
\end{figure}

\begin{figure}[h]
    \centering
    \begin{subfigure}[b]{0.45\textwidth}
        \centering
        \includegraphics[width=\columnwidth, trim=0cm 5cm 0cm 14cm]{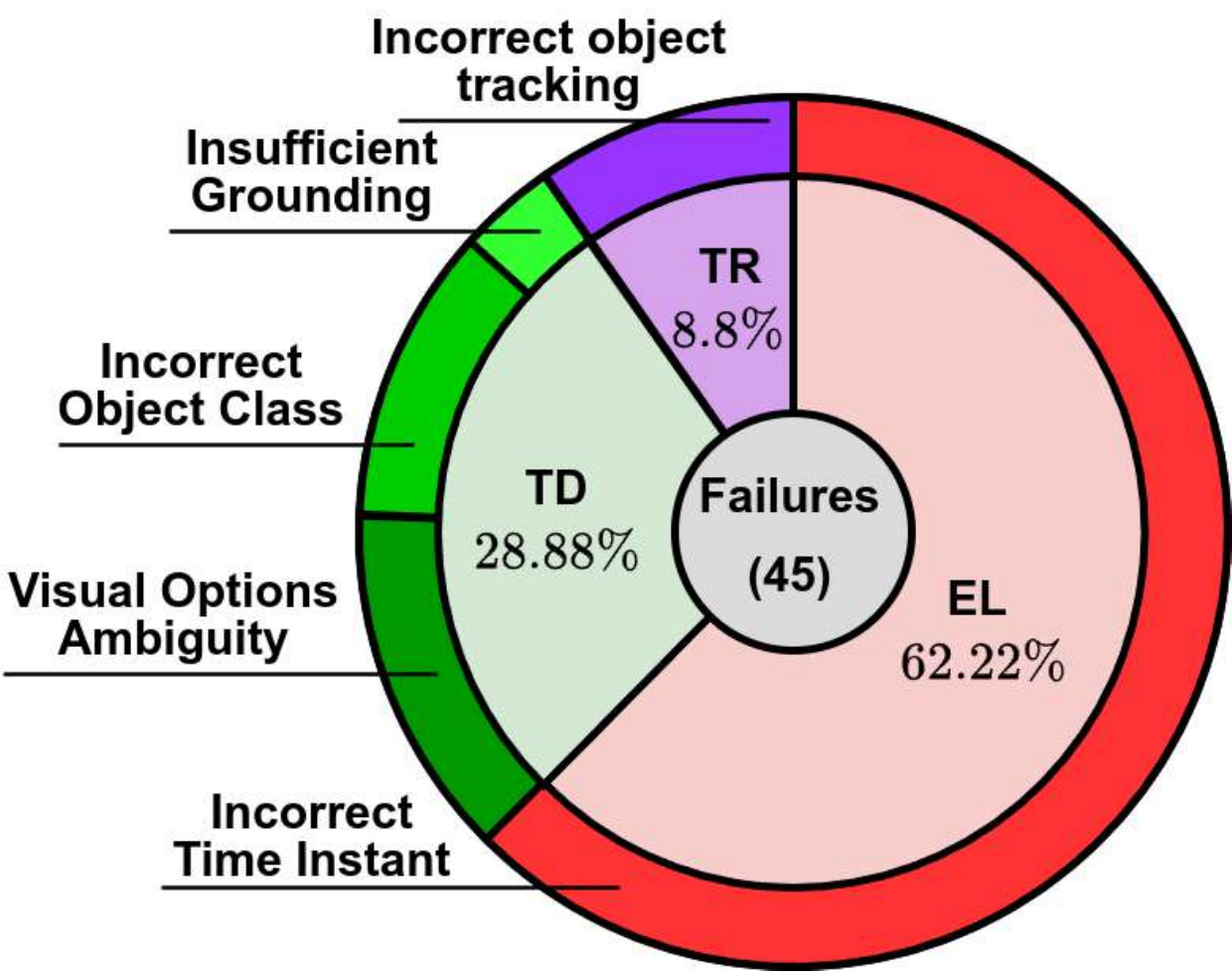} 
        \caption{Component-Wise Analysis}
        \label{fig:subfig1}
    \end{subfigure}
    \hfill
    \begin{subfigure}[b]{0.45\textwidth}
        \centering
        \includegraphics[width=\columnwidth, trim=0cm 5cm 0cm 8cm]{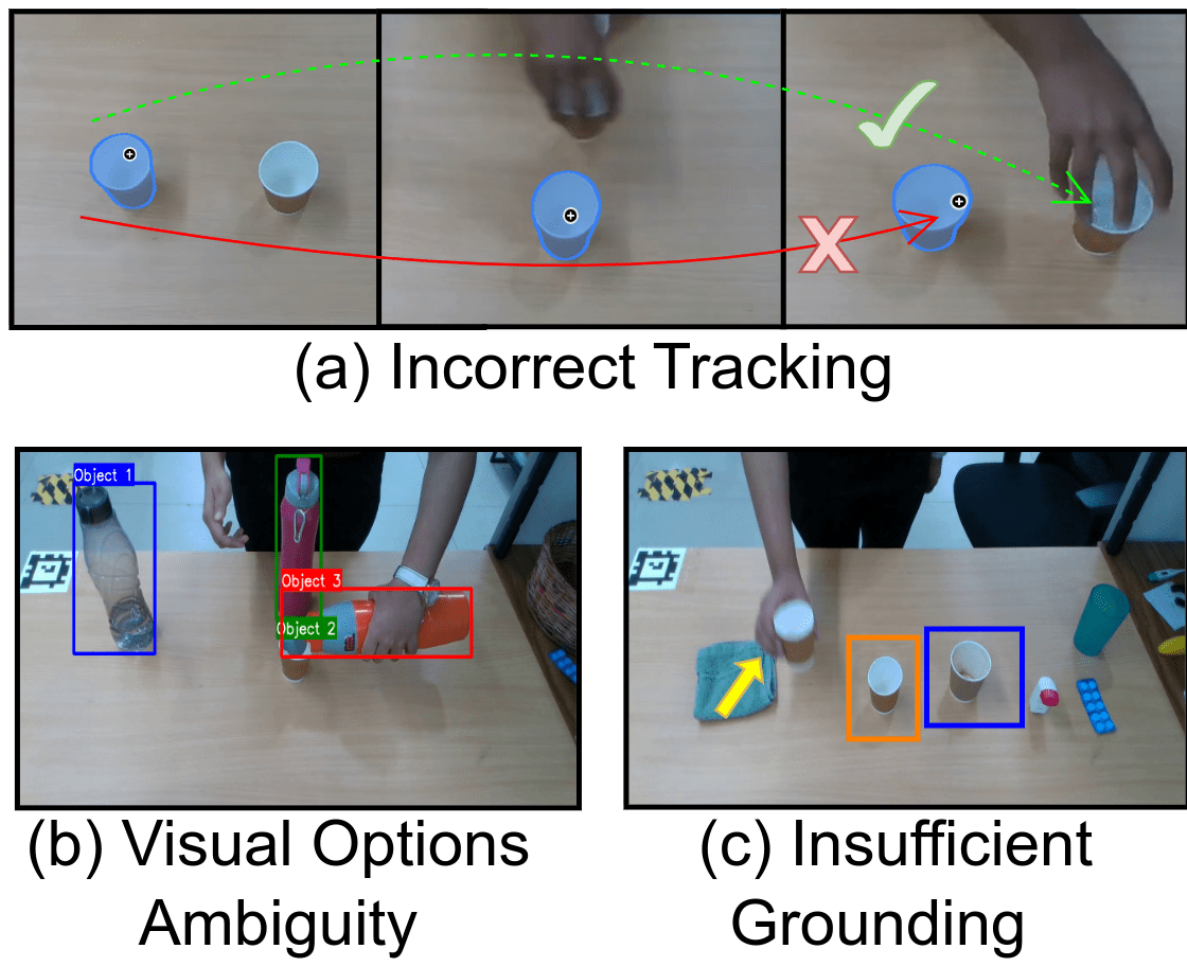} 
        \caption{Examples of Failure Cases}
        \label{fig:subfig2}
    \end{subfigure}
    \caption{\textbf{(a)} Frequency of failure cases generated by each module. \textbf{(b)} Instances of failure cases in our pipeline.}
    \label{fig:component-failure}
\end{figure}

\subsection{Analysis of Model Components}
We evaluated multiple state-of-the-art models on our dataset across all tasks (see results in Figure \ref{fig:model_analysis}). Training video-understanding models is constrained by the limited amount and variety of available data, as high-quality video question answering (QA) annotation is expensive. CogVLM2-Video \cite{hong2024cogvlm2} introduces an automated process for 
generating a post-training dataset of 30k video-QA samples, improving candidate-interval extraction. While approaches like \cite{videollava} and \cite{videochat2} compress video content by sampling images and captioning them, they lose temporal awareness, hindering precise frame-timestamp associations. In contrast, CogVLM2-Video retains temporal reasoning by annotating frames with timestamps for better performance. For the Event Localization task, the ground truth was a manually annotated timestamp. 
For the Temporal Reasoning task, the outputs were manually evaluated and all sufficient answers were considered correct, even if they were not strictly necessary. All of the models in Figure \ref{fig:model_analysis} were evaluated assuming that their inputs were correct.

\section{Conclusion}
In this paper, we present G\textsuperscript{2}TR, a novel approach to grounded temporal reasoning. We factorize the problem into three key components: (i) candidate interval localization in a video based on required interactions, (ii) 
fine-grained spatial reasoning within the localized interval to ground the target object, and (iii) tracking the object post-interaction. By leveraging pre-trained visual language models and large language models, G\textsuperscript{2}TR achieves zero-shot generalization for both object set and interactions. We also propose a dataset of 155 video-instruction pairs covering spatially complex, multi-hop, partially observable, and multi-interaction temporal reasoning tasks. Evaluation on the dataset shows significant improvement over alternative approaches, highlighting G\textsuperscript{2}TR’s potential in robot instruction following. Finally, it is important to note that G\textsuperscript{2}TR currently has two limits: (i) it can only process videos up to one minute, as constrained by the video-reasoning model, and (ii) it can ground only a single object at a time. We aim to overcome both these limitations in future.




\newpage

\newpage
\appendix

\end{document}